\documentclass[nodate]{article}

\usepackage{arxiv}

\usepackage{hyperref}
\usepackage{wrapfig}
\usepackage{amsmath, amssymb, bm}
\usepackage{cleveref}
\usepackage{paralist}
\usepackage{xspace}
\usepackage{booktabs}
\usepackage{amsfonts}
\usepackage{nicefrac}
\usepackage{microtype}
\usepackage{fancybox}
\usepackage{makecell}
\usepackage{xcolor}
\usepackage{graphicx}
\usepackage[utf8]{inputenc}
\usepackage[T1]{fontenc}
\usepackage{url}
\usepackage{lipsum}
\usepackage{natbib}
\usepackage{tcolorbox}
\usepackage{siunitx}
\usepackage{tabularx}

\usepackage{scalerel,xparse} 

\NewDocumentCommand{\brideemoji}{}{%
  \protect\raisebox{-0.2ex}{%
    \protect\includegraphics[height=0.9em]{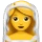}%
  }%
}
\NewDocumentCommand{\groomemoji}{}{%
  \protect\raisebox{-0.2ex}{%
    \protect\includegraphics[height=0.9em]{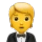}%
  }%
}

\NewDocumentCommand{\pigemoji}{}{%
  \protect\raisebox{-0.2ex}{%
    \protect\includegraphics[height=0.9em]{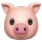}%
  }%
}

\graphicspath{{figures/}}

\title{Improving Steering Vectors by\\Targeting Sparse Autoencoder Features}
\author{
    Sviatoslav Chalnev$^*$ \\
    \texttt{slava.chalnev@gmail.com} \\
    \And
    Matthew Siu$^*$ \\
    \texttt{matthewwilsonsiu@gmail.com} \\
    \AND
    Arthur Conmy \\
    \texttt{arthurconmy@gmail.com} \\
}

\date{}
\begin{document}
\maketitle

\renewcommand{\thefootnote}{\fnsymbol{footnote}}
\footnotetext[1]{Equal contribution}
\footnotetext[2]{Code available at \url{https://github.com/slavachalnev/SAE-TS}}

\begin{abstract}
To control the behavior of language models, steering methods attempt to ensure that outputs of the model satisfy specific pre-defined properties. Adding steering vectors to the model is a promising method of model control that is easier than finetuning, and may be more robust than prompting.However, it can be difficult to anticipate the effects of steering vectors produced by methods such as CAA \citep{panickssery2024steeringllama2contrastive} or the direct use of SAE latents \citep{templeton2024scaling}. In our work, we address this issue by using SAEs to measure the effects of steering vectors, giving us a method that can be used to understand the causal effect of any steering vector intervention. We use this method for measuring causal effects to develop an improved steering method, \textbf{SAE-Targeted Steering} (SAE-TS), which finds steering vectors to target specific SAE features while minimizing unintended side effects. We show that overall, SAE-TS balances steering effects with coherence better than CAA and SAE feature steering, when evaluated on a range of tasks.
\end{abstract}


\section{Introduction}
\label{sec:introduction}

There are widespread calls for better control of the behaviour of Large Language Models (LLMs; e.g. \citet{whitehouse2023ai}). Current methods such as prompting \citep{wallace2024instructionhierarchytrainingllms} and finetuning \citep{ouyang2022traininglanguagemodelsfollow, chung} offer some degree of control, but have clear limitations. For example, prompting can be fragile and is often susceptible to methods that can subvert these instructions \citep{wei2023jailbrokendoesllmsafety}. Finetuning a model can be more robust but requires a curated dataset for training which can be both expensive and time-consuming to produce (e.g. \citet{dubey2024llama3herdmodels}).

\textit{Steering vectors} \citep{turner2024activationadditionsteeringlanguage} have the potential to be more robust than prompting, and both cheaper and easier to implement than finetuning. Steering vectors work by adding activations to the hidden state of a model, part way through the forward pass (\Cref{sec:measuring steering effects}). By generating and inserting activation vectors into the model's forward pass, we can steer the model towards desired behaviors.

However, a problem with current steering methods is their unpredictability -- it's often unclear exactly how a steering vector will affect model behavior. Steering vectors may not produce the intended changes in the model's output or may cause unforeseen behaviors, as we discuss in \Cref{sec:measuring steering effects}. In some cases, steering interventions produce no interpretable changes in model behavior other than model degradation, as we show in \Cref{sec:eval}. This unpredictability makes it difficult to precisely control the model.

\begin{figure}[htbp]
  \centering
  \hspace{-0.1cm}
  \includegraphics[width=0.89\linewidth]{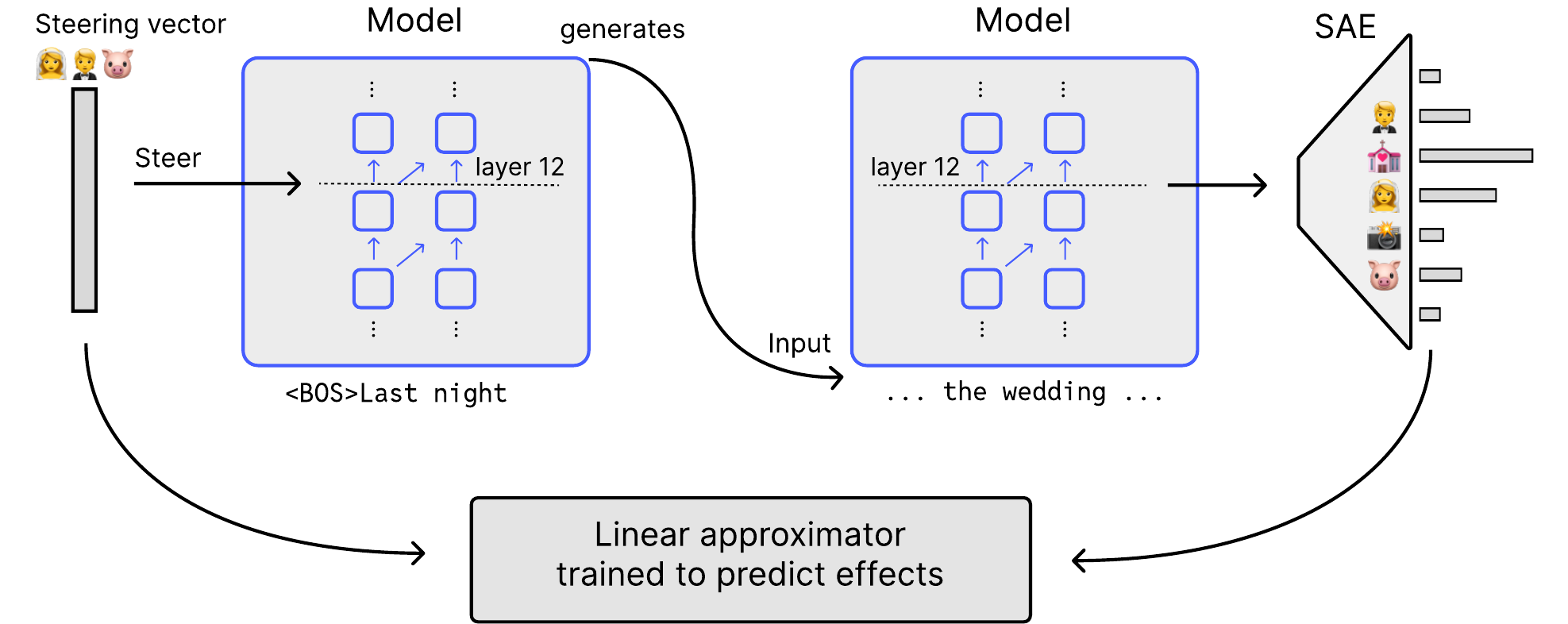}
  \caption{Diagram showing the feature effects measurement method. We sample a steering vector (depicted here as \brideemoji\groomemoji\pigemoji \space to emphasise that it represents a mixture of concepts), and steer the model by adding it to the residual stream activations at layer 12. We use this steered model to generate text completions starting from a prompt (e.g. "<BOS>Last night"). These completions are then fed back through the model up to layer 12, where the activations are passed through a Sparse Autoencoder (SAE) and averaged to measure feature effects. The linear approximator is trained on pairs of steering vectors and their measured feature effects to predict effects for new steering vectors.}
  \label{fig:full_diagram}
\end{figure}

To address these challenges, we develop a method for quantifying the effects of a steering intervention on model outputs. We use Sparse Autoencoders (SAEs; \citet{ng, cunningham2023sparseautoencodershighlyinterpretable, bricken2023monosemanticity}) to measure the change in feature activations caused by steering interventions. This lets us understand what change in behavior to expect from any particular steering intervention. Note: our analysis is \textit{not} about how early SAEs impact later layer SAEs (as studied by prior work such as \citet{marks2024feature} and \citet{anders_bloom_2024_gpt2saeacts}). Instead, we look at how rollouts generated by a steered model differ from those generated by the unsteered, base model.

Building on this feature effects measurement method, we introduce \textbf{SAE-Targeted Steering} (SAE-TS), a method that constructs steering vectors to specifically target desired SAE features while minimizing unintended side effects. SAE-TS involves learning a linear relationship between steering vectors and their effects on SAE features. This allows us to construct steering vectors using the interpretable features found by the SAE. Notably, SAE-TS uses the SAE for steering differently to prior work such as \citet{templeton2024scaling} and \citet{durmus2024steering}. We use the SAE to measure the effects of steering vectors, rather than directly adding SAE latents to model forward passes.

We evaluate SAE-TS against existing methods such as Contrastive Activation Addition (CAA) \citep{panickssery2024steeringllama2contrastive} and direct SAE feature steering \citet{templeton2024scaling}. Our evaluations demonstrate that SAE-TS outperforms existing methods, achieving better alignment with the intended behavior while maintaining the semantic coherence of the generated text across various tasks.

\textbf{Key Contributions:}
\begin{compactenum}
    \item We develop a method to quantify and interpret the effects of a steering intervention using SAEs (\Cref{sec:measuring steering effects}).
    \item We introduce SAE-Targeted Steering (SAE-TS), a method that constructs steering vectors to achieve specific desired effects while minimizing unintended changes (\Cref{sec:optimised_steering_vectors}).
    \item We evaluate SAE-TS on a set of steering tasks (\Cref{sec:eval}), showing that it outperforms existing methods, successfully steering the model while maintaining output quality.
\end{compactenum}

\section{Related work}
\label{sec:related work}

\paragraph{Activation Steering} is a method for controlling model outputs introduced in \citet{turner2024activationadditionsteeringlanguage} where steering vectors are extracted by taking the difference between activations from pairs of contrasting positive and negative prompts. \citet{panickssery2024steeringllama2contrastive} scaled this method to Llama-2, and \citet{repe} use the methodology on a somewhat wider range of tasks. Recently, \citet{cao2024personalizedsteeringlargelanguage} offers a way to learn a steering vector to optimally steer a model toward producing desired outputs. As with the activation steering methods above, a dataset of positive and negative examples is required for this method to work. \citet{lee2024programmingrefusalconditionalactivation} introduce conditional steering. \citet{causalSteering} also studies the causal limitations of existing steering methods.

\textbf{Mechanistic Interpretability and SAEs.} Mechanistic Intepretability aims to understand how LLMs function internally by breaking down the models into understandable components \citep{olah2020zoom, AnthropicMechanisticEssay}. The recent development of \textbf{sparse autoencoders} \citep{ng, bricken2023monosemanticity, cunningham2023sparseautoencodershighlyinterpretable} gives us a way to decompose the residual stream of transformer models into sparse and often human understandable features. Sparse autoencoders provide evidence supporting the hypothesis that the latent space of LLMs is composed of linear and interpretable directions \citep{arora, elhage2022toy}. \citet{templeton2024scaling} and \citet{zhao2024steeringknowledgeselectionbehaviours} demonstrate that these directions found by the SAEs can be used for steering.

\section{Measuring Steering Effects}
\label{sec:measuring steering effects}

As discussed in the introduction (\Cref{sec:introduction}), it can be difficult to predict the behavior of steering vectors. Measuring the effects of a steering intervention is useful for both interpreting the causal role of features and designing more effective steering vectors.

In transformer models, we can represent the forward pass as $h_n = b_n(h_{n-1})$, where $b_i$ denotes the $i^\text{th}$ block and $h_i$ is the hidden state after layer $i$. When we insert a steering vector $\bm{v}$ at a specific layer $l$, we modify the hidden state as $h_l \leftarrow h_l + \alpha \bm{v}$, where $\alpha$ is a scaling coefficient that determines the norm of the added steering vector.

We use JumpReLU SAEs \citep{rajamanoharan2024jumpingaheadimprovingreconstruction}, which consist of an encoder $f$ and a decoder, where the reconstruction $\hat{\bm{x}}$ of the input $\bm{x}$ is given by:
\begin{equation}
    \hat{\bm{x}} = f(\bm{x}) \bm{W}_{\mathrm{dec}} + \bm{b}_{\mathrm{dec}},
\end{equation}
and the SAE is trained such that $\hat{\bm{x}}$ approximates $\bm{x}$. SAE steering involves using the decoder vector $d_i$ of feature $i$, which is the $i^\text{th}$ row of $\bm{W}_{\mathrm{dec}}$, as the steering vector.

In order to measure a steering intervention's effect on the model output, we look at the difference in SAE feature activations between steered model outputs and original, unsteered model outputs. Our method is outlined at a high level in \Cref{fig:full_diagram} and can be summarised as follows:

\begin{enumerate}
    \item \textbf{Data Generation}: Generate text rollout datasets $D_{\mathrm{steered}}$ and $D_{\mathrm{unsteered}}$ using steered and unsteered models respectively, where the steered model has a steering vector $\bm{v}$ inserted at layer $l$.
    \item \textbf{Feature Extraction}: For each generated output, pass the text back through the model up to layer $l$, and use the SAE encoder $f$ to extract the feature activations at each position.
    \item \textbf{Effect Computation}: Compute the average feature activations across all outputs and positions for both the unsteered and steered models, and calculate the difference to obtain the \emph{steering effects} vector $\bm{y} \in \mathbb{R}^{d_\mathrm{SAE}}$:
    \begin{equation}
        \bm{y} = \mathbb{E}_{\text{steered}}[f(\bm{x})] - \mathbb{E}_{\text{unsteered}}[f(\bm{x})].
    \end{equation}
\end{enumerate}

We use the Gemma-2-2B model \citep{gemmateam2024gemma2improvingopen} and a Gemma Scope layer 12 residual stream SAE with hidden dimension of 16,384 \citep{lieberum2024gemmascopeopensparse}. We run the above procedure for every feature of the SAE, generating 896 rollouts of length 32, and computing the average difference between the steered and unsteered feature activations.

\Cref{tab:feature_effects_months} shows a cherry-picked example of a feature which, when used for steering, has feature effects that are not itself. This feature (id 12904) activates on months (e.g. "January") but steering with its decoder vector causes the model to output digits of years. Upon closer inspection we see that the feature only activates on months following a day (e.g. 1st of January), so it is unsurprising that steering with this feature causes the model to output years. Some of the largest feature effects are features which are not clearly interpretable but are instead commonly occurring. E.g. in \Cref{tab:feature_effects_months} the top effect is feature id 6810 with an activation difference more than 3 times higher than the next highest activation difference.

\begin{table}[h!]
\centering
\begin{tabularx}{0.9\textwidth}{|c|c|X|}
\hline
\textbf{Act Diff} & \textbf{Feature id} & \textbf{Feature description} \\
\hline
4.5819 & 6810  & No clear pattern. Commonly occurring feature with 45\% activation density \\
\hline
1.3763 & 8641  & The number two. \\
\hline
1.3080 & 8223  & Space before a number. \\
\hline
0.9797 & 8233  & Last digit of a year. \\
\hline
0.9769 & 10356 & Third digit of a 2000s year e.g. 20\textbf{1}5. \\
\hline
0.7754 & 847   & The zero digit. \\
\hline
0.7542 & 14914 & The digit following a two. \\
\hline
0.7470 & 7517  & No clear pattern. Commonly occurring feature with 22\% activation density \\
\hline
0.7069 & 2381  & Zero following a two. \\
\hline
0.6341 & 8258  & Second digit of a year, mostly 9. \\
\hline
\end{tabularx}
\vspace{1mm}
\caption{Table showing activations, feature ids, and feature descriptions of the SAE feature steering vector (ft id 12904) in Gemma-2-2b. This feature activates on months following a day (e.g. "1st of \textbf{January}").}
\label{tab:feature_effects_months}
\end{table}

Given our initial motivation to use steering effects to develop better causal explanations of a feature's \textit{role} in the model, features such as the month feature, which only have significant effects on features other than themselves, are of particular interest. When inspecting these features, we find that the effects encode many reasonable grammatical and semantic relationships. Another example is a feature that activates on "say", but increases a feature that activates on "hello" (see \Cref{sec:feature_effect_examples}).

\section{Targeted Steering}
\label{sec:optimised_steering_vectors}
We can use the above method of measuring steering effects to find steering vectors which steer the model towards a desired feature effect, while having minimal side-effects.

We do this by first training a linear \textbf{effect approximator} function (see \Cref{sec:effect_approximator}) to predict the feature effects for any given steering vector and then use this function to find targeted steering vectors to achieve a desired effect.

The effect approximator is a linear function $\bm{\hat{y}} = \bm{x}\bm{M} + \bm{b}$, where $\bm{x}$ is the $d_{\mathrm{model}}$-dimensional steering vector, $\bm{M}$ is a $d_{\mathrm{model}} \times d_{\mathrm{sae}}$ matrix, and $\bm{b}$ has dimension $d_{\mathrm{sae}}$. We train it to minimize the mean squared error (MSE) between its prediction $\bm{\hat{y}}$ and the observed effect vector $\bm{y}$.

Once we have trained the effect approximator, we use it to find a steering vector that increases the activation of a target feature $j$, while keeping other features unchanged. To achieve this, we construct the targeted steering vector $\bm{s}$ as follows:
\begin{equation}
    \bm{s} = \frac{\bm{M}_j}{\|\bm{M}_j\|} - \lambda \frac{\bm{M}\bm{b}}{\|\bm{M}\bm{b}\|}
    \label{eq:targeted_steering}
\end{equation}
We then normalize $\bm{s}$ to unit norm. In all our experiments, we set $\lambda = 1$. See \Cref{sec:why not inverse} for further discussion on why we do this instead of solving for $\bm{x}$ by computing the pseudoinverse of $\bm{M}$.

\subsection{Training the Effect Approximator}
\label{sec:effect_approximator}

To construct the effect approximator, we collect a dataset of 50,000 steering vectors and their corresponding steering effects. The steering vectors are the decoder vectors from a larger SAE trained at the same layer, which in the case of Gemma-2-2b is the layer 12 65k SAE with an L0 of 72. This provides us with a diverse set of steering vectors, letting us observe a wide range of steering effects.

For each steering vector $\bm{x}$ in the dataset, we measure its effect $\bm{y}$ on the model by computing the difference in SAE feature activations between the steered and unsteered model outputs, as described in Section~\ref{sec:measuring steering effects}. We then train the linear effect approximator $\hat{\bm{y}} = \bm{x}\bm{M} + \bm{b}$ using Adam to minimize the MSE between $\hat{\bm{y}}$ and $\bm{y}$.

By learning the mapping from steering vectors to their effects on SAE features, the effect approximator allows us to predict the change in feature activations for any given steering vector. This enables us to design steering vectors that produce precise and predictable changes in the model's behavior.

\subsection{Scaling Factor Selection}
\label{sec:scaling_factor_choice}

Selecting an appropriate scaling factor $\alpha$ when adding the steering vector $\bm{x}$ is important for getting good steering results. The model is sensitive to different directions to varying degrees; some steering vectors may have a significant impact even with a small scaling factor, while others require a larger scaling factor to produce noticeable effects. To compensate for this variability, we need an automatic method to adjust the scaling factor for each steering vector individually.

We do this by finding a scaling factor $\alpha$ for each steering vector such that the steered model’s cross-entropy loss goes up by 0.5 above the unsteered baseline. By applying this scaling factor selection process to all steering vectors in our dataset, we ensure that the collected steering effects accurately reflect the model's response within the optimal range of steering intensities.

\section{Evaluations}
\label{sec:eval}
To evaluate our steering vectors, we use gpt-4o-mini to rate two aspects of the generated text on a scale of 1 to 10:
\begin{itemize}
    \item \textbf{Behavioral score} Assesses whether the steering target was met, based on task-specific criteria.
    \item \textbf{Coherence score} Evaluates whether the steering intervention maintains the model's general capabilities and produces semantically correct text.
\end{itemize}

The ratings are averaged and normalised to give scores between 0 and 1. We multiply these two scores to obtain the \textbf{Behavioral*Coherence score}, which is the main metric we use for measuring steering quality. This way, our evaluations require both achieving the desired steering effect and maintaining the overall quality of the generated text.

\begin{figure}[htbp]
  \centering
  \includegraphics[width=\linewidth]{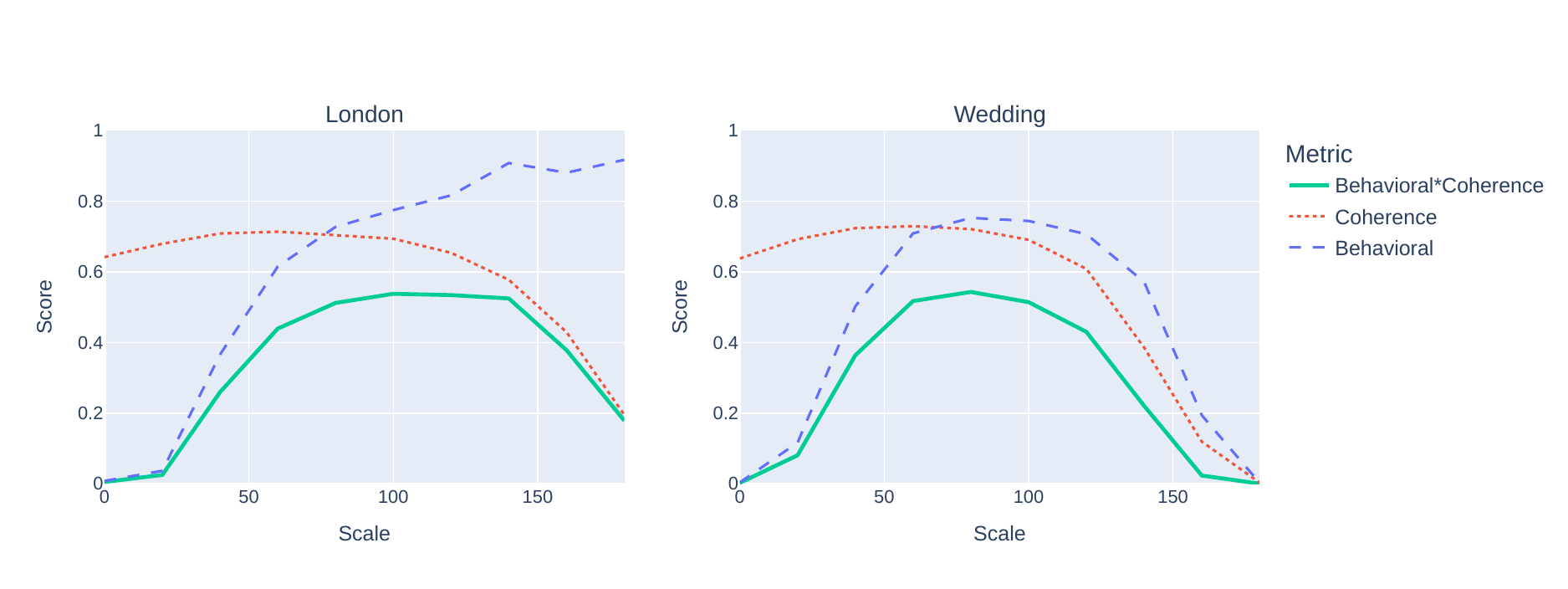}
  \caption{Plots showing Behavioral, Coherence, and Behavioral*Coherence scores for the London and Wedding tasks at varying steering scales.}
  \label{fig:LondonWedding}
\end{figure}

We focus on steering a base language model (Gemma-2-2b) and evaluate on open-ended generation starting from the prompt "<BOS>I think". The task is to steer the model so that it discusses a pre-specified topic, while remaining coherent. We selected 9 topics, most of which appear in prior steering work. We steer the model by adding the steering vector at some scale $\alpha$ to the residual stream at every token position. We sample 256 steered text completions, each 32 tokens long.

Steering is sensitive to the norm of the steering vector so we vary the scale $\alpha$ and plot the three scores for our SAE-TS steering method as shown in \Cref{fig:LondonWedding}. Starting from scale $\alpha=0$ (the unsteered model), the Coherence score is 0.64 while the Behavioral score is 0, making the Behavioral*Coherence score 0. As we increase the scale, the Behavioral score either keeps rising to more than 0.9 in the case of the "London" task, or peaks at around 0.75 and then goes back down as is the case in the "Wedding" task. Interestingly, the Coherence score first rises slightly to a peak of around 0.71 and then drops off. This rising Coherence score phenomenon is specific to the SAE-TS method and we explore it further in \Cref{sec:importance_of_bias}.

We benchmark three steering methods, comparing their Behavioral*Coherence scores across steering vector scales:
\begin{itemize}
    \item \textbf{Contrastive Activation Addition} (CAA), which we define as the mean difference of model activations between a set of positive and negative prompts, averaged over token positions and prompts. This is similar but not identical to \citet{panickssery2024steeringllama2contrastive}'s usage.
    \item \textbf{SAE feature steering}, using the decoder vector of the relevant SAE feature.
    \item \textbf{SAE targeted steering} (SAE-TS), as described in \Cref{sec:optimised_steering_vectors}.
\end{itemize}

\begin{figure}[htbp]
  \centering
  \includegraphics[width=\linewidth]{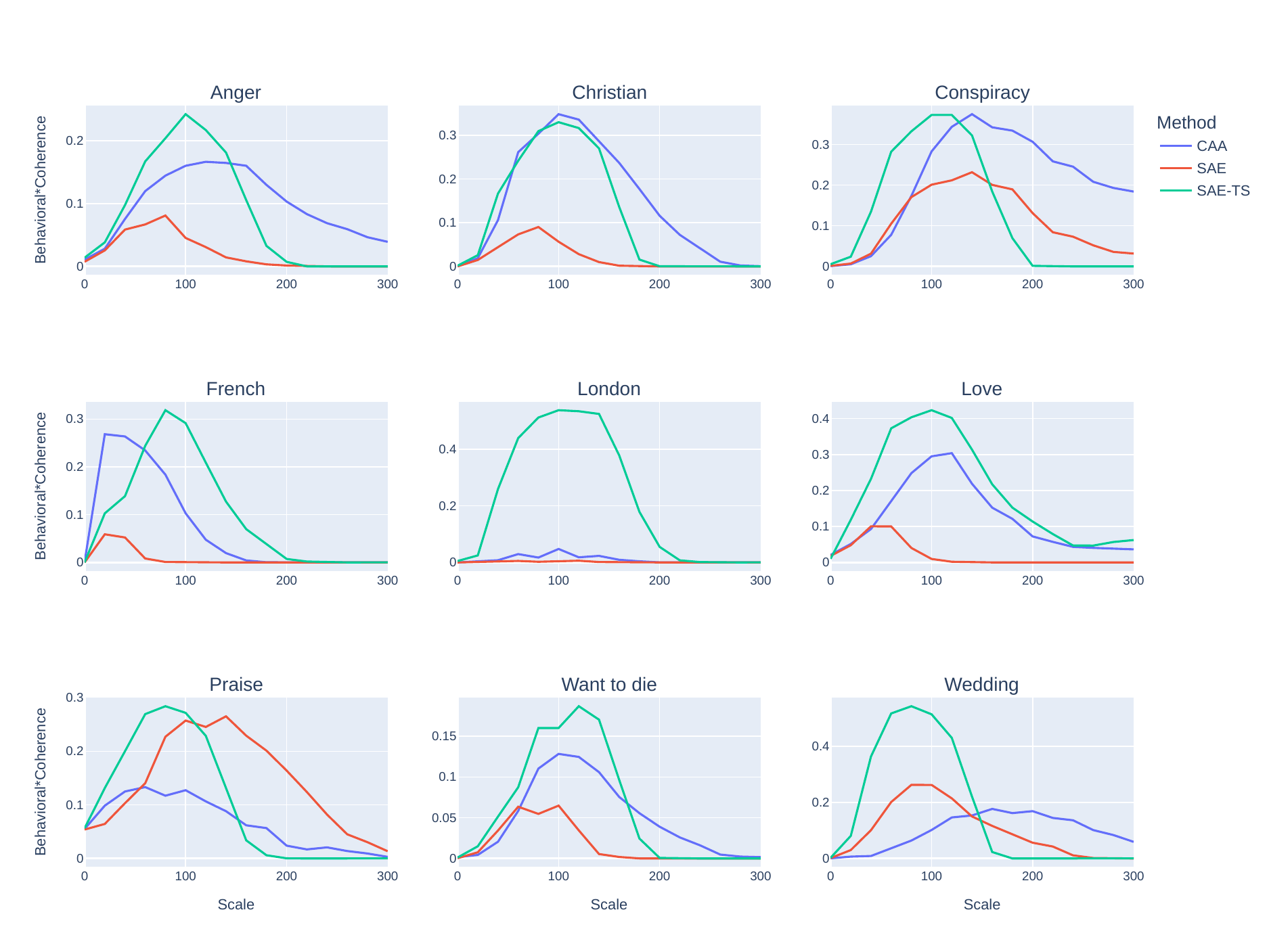}
  \caption{Plots showing Behavioral*Coherence score for the CAA, SAE, and SAE-TS steering methods on all 9 tasks. We see that SAE-TS is superior to the other two methods on 7 of the 9 tasks across a wide range of steering scales.}
  \label{fig:all_plots}
\end{figure}

\Cref{fig:all_plots} shows the performance of these methods across the 9 steering tasks. The $x$-axis represents the scaling factor $\alpha$ while the $y$-axis is Behavioral*Coherence score. Looking at the peak of the score (how good is it when the appropriate scale is chosen), displayed in \Cref{tab:steering_comparison}, we see that the our SAE-TS method outperforms the baseline methods on 7 of the 9 tasks. The two exceptions are the "Christian" task, where CAA is the best method, and the "Conspiracy" task, where CAA is tied with SAE-TS.

On the "London" task, CAA and SAE steering perform unusually poorly, while SAE-TS performs very well. Even at high steering scales, CAA and SAE steering do not produce mentions of London. Although SAE-TS is significantly better than the other two methods for this task, we do observe an interesting failure mode of SAE-TS at high scaling factors ($\alpha > 160$, well above optimal scale). At these scales, the output diversity collapses and the model begins to produce text focused entirely on fashion shows and art exhibitions in London, as we show in example outputs in \Cref{sec:example_rollouts}.

While we primarily focus on Gemma-2-2B in this paper, we also validate our methods on Gemma-2-9B and present these results in \Cref{sec:9B_results}. We observe broadly similar trends, with SAE-TS outperforming the baseline methods on average, though we see that CAA sometimes performs better on the larger model.

\begin{table}[h!]
\centering
\begin{tabular}{lccc}
\hline
\textbf{Steering Goal} & \textbf{ActSteer} & \textbf{SAE} & \textbf{SAE-TS (ours)} \\
\hline
Anger          & 0.1666 & 0.0810 & \textbf{0.2424} \\
Christian      & \textbf{0.3486} & 0.0902 & 0.3302 \\
Conspiracy     & \textbf{0.3748} & 0.2319 & 0.3729 \\
French         & 0.2684 & 0.0589 & \textbf{0.3186} \\
London         & 0.0476 & 0.0061 & \textbf{0.5380} \\
Love           & 0.3039 & 0.1004 & \textbf{0.4234} \\
Praise         & 0.1332 & 0.2654 & \textbf{0.2842} \\
Want to die    & 0.1284 & 0.0649 & \textbf{0.1870} \\
Wedding        & 0.1768 & 0.2626 & \textbf{0.5432} \\
\hline
Average        & 0.2165 & 0.1290 & \textbf{0.3600} \\
\hline
\end{tabular}
\caption{Table showing maximum steering scores.}
\label{tab:steering_comparison}
\end{table}

\section{Feature Effects Visualization}

In this section, we introduce EffectVis, an interactive interface and tool built for exploring feature effects. On the top left panel, users can search for features and view their feature visualization charts – similar to Neuronpedia \citep{Neuronpedia}. Features can be added as feature cards to the canvas when you want to inspect them further.

\begin{figure}[htbp]
  \centering
  \includegraphics[width=\linewidth]{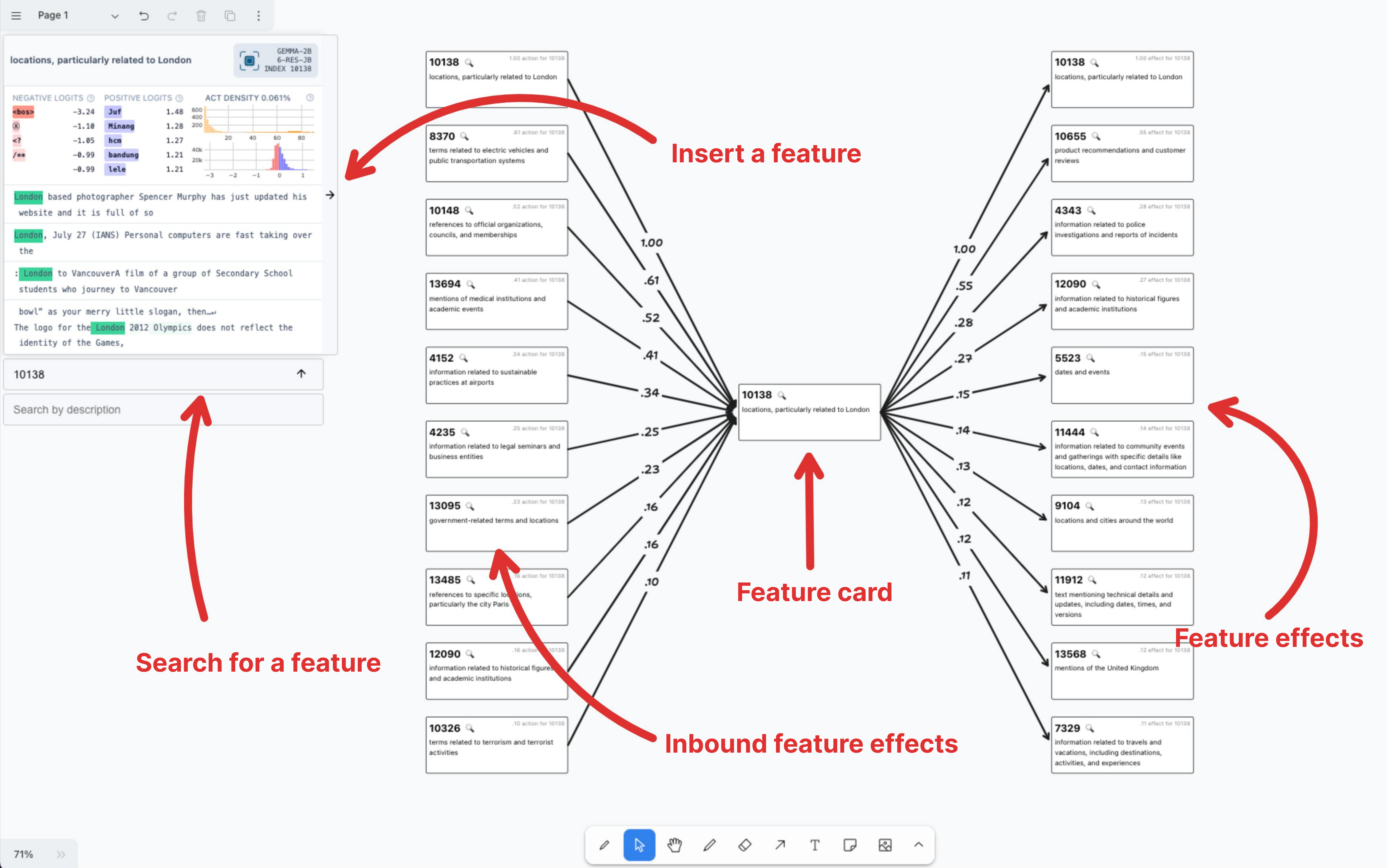}
  \caption{Interface overview for EffectVis. Available at \href{https://effectvis.vercel.app/}{https://effectvis.vercel.app/}}
  \label{fig:interface_overview}
\end{figure}

For a selected card, our tool allows you to query for three different properties of the feature:
\begin{itemize}
    \item The feature effects.
    \item The feature actions. These are the features that cause the selected feature to activate more often.
    \item Cosine similar feature directions.
\end{itemize}

In addition, when a feature card is selected, other cards for the same feature are highlighted, making it easy to see here else a feature is used at a glance. By clicking the magnifying icon on the top right of a feature card, the top panel will be populated with the relevant information for that feature.

Prior to developing EffectVis, we had a multi-step process for exploring and making sense a feature's effects and/or cosine similarity. First, we would access the list of effects for a feature in a python notebook. Then, using Neuronpedia, we would search for each feature one by one. These were then added to a list on Neuronpedia.

Due to the time-consuming nature of the process, we were limited in how many features we could inspect, reducing our ability to build intuition for what was being measured with the feature effects. EffectVis solved this for us by compressing the above steps into a single keyboard command action. It also made comparisons between feature effects and/or decoder cosine similarity easier. Lists of features could be pulled up side-by-side and compared visually. For comparing many lists, selecting a card shows you the other lists it is a part of.

EffectVis helped us develop the insights that led to our Optimized Steering Vectors method. In the future, it may useful to incorporate functionality for querying our feature effects by different filters such as effect scores as we did for part of our analysis in the feature effects section. This would allow us to perform more experiments and ask a wider variety of questions using the interface that we resorted to using python notebooks to explore. We haven't rigorously compared EffectVis to alternative tools, as it is tailored for the standalone method that we developed.

\section{Discussion}
\subsection{Limitations}
We performed all of our investigations using Gemma-2 2B and 9B, both of these models share the same distilled model architecture, so our results may not transfer to other models. Additionally, these are base models which haven't been instruction tuned, so we focused on steering in the open-ended generation setting and did not investigate tasks and features which would be relevant for safety or capabilities.

We focused only on steering towards concepts present in SAEs, which is a problem for multiple reasons. Firstly, some concepts don't appear as individual SAE features, a phenomenon called feature splitting. Secondly, niche features may only appear in extremely large SAEs, as discussed in the "feature completeness" section of \citet{templeton2024scaling}. These problems are particularly relevant because our targeted steering method, as described in \Cref{sec:optimised_steering_vectors}, can only steer towards features present in the SAE used to measure feature effects. We explore a potential solution in \Cref{sec:rotation_steer}.

\subsection{Future work}
An important direction for future work is to apply our steering methods to chat models and safety-relevant steering targets to evaluate effectiveness in practical scenarios. Additionally, this work can be extended by exploring different SAE architectures, such as TopK SAEs \citep{gao2024scalingevaluatingsparseautoencoders}, and testing on various model architectures beyond the Gemma family. Another direction to explore is adapting BiPO \citep{cao2024personalizedsteeringlargelanguage} to optimize steering vectors towards desired feature effects.

\section*{Acknowledgements}
This research was conducted as part of the MATS 6.0 program. We would like to thank Yeu-Tong Lau and Kai Fronsdal for valuable discussions that helped develop the SAE-TS method. We would also like to thank the GitHub user \url{https://github.com/spfaul} for finding a bug in our CAA implementation which we have fixed and regenerated results for, in this version of the paper.

\bibliographystyle{plainnat}
\bibliography{references}  

\clearpage
\appendix
\section*{Appendix}

\section{Targeted Steering Method Justification}
\label{sec:why not inverse}

\subsection{Empirical justification}
As described in \Cref{sec:optimised_steering_vectors}, our targeted steering method aims to find a steering vector $\bm{s}$ that increases activation of a target feature while minimizing unintended effects. Given our effect approximator function $\hat{\bm{y}} = \bm{x}\bm{M} + \bm{b}$, an obvious approach would be to directly solve for the optimal steering vector $\bm{x}$ by computing the pseudoinverse of $\bm{M}$. However, empirically we found that our method (SAE-TS), which takes the normalized target feature column of $\bm{M}$ and subtracts the normalized bias effects, significantly outperforms the pseudoinverse approach, as shown in \Cref{tab:pseudoinverse_vs_ours}.

\begin{table}[h!]
\centering
\begin{tabular}{lcc}
\hline
\textbf{Steering Goal} & \textbf{SAE-TS} & \textbf{Pseudoinverse} \\
\hline
Anger & \textbf{0.2424} & 0.1899 \\
Christian & \textbf{0.3302} & 0.0025 \\
Conspiracy & \textbf{0.3729} & 0.0644 \\
French & \textbf{0.3186} & 0.1912 \\
London & \textbf{0.5380} & 0.0722 \\
Love & \textbf{0.4234} & 0.2682 \\
Praise & \textbf{0.2842} & 0.0638 \\
Want to die & \textbf{0.1870} & 0.1683 \\
Wedding & \textbf{0.5432} & 0.0567 \\
\hline
\end{tabular}
\caption{Table showing maximum steering scores for our SAE-TS method vs computing the optimal steering vector using the pseudoinverse.}
\label{tab:pseudoinverse_vs_ours}
\end{table}

\subsection{Why do this work? Intuition}
If we ignore the bias term, using the normalized $\bm{M}_j$ is equivalent to linear regression and gives the steering direction that maximizes predicted effect on feature $j$ subject to the unit norm constraint, while ignoring effects on other features. Why does ignoring predicted effects on other features produce good results?

An important part of the story is the data on which the effect approximator was trained. The steering vectors used to generate the data were added at a scale $\alpha$ chosen to increase the model's loss by 0.5. This scaling encodes information about the model's sensitivity to different directions - directions that strongly affect many features require a smaller $\alpha$ to reach the target loss increase.

Then, when training the effect approximator, the steering vectors are normalised to have norm 1. As a result, $\bm{M}_j$ implicitly accounts for effects on other features. A steering direction that strongly affects other features will show up in the training data with smaller effect magnitudes due to the smaller $\alpha$ used.

\subsection{Effect of the bias term}
\label{sec:importance_of_bias}

The effect approximator's bias vector $\bm{b}$ has a few features with large values while most are close to 0. The high-bias features include three that activate on the <BOS> token, an induction feature that activates on repeated patterns, and a position feature that activates on all token positions > 2. All the largest features are densely activating, as shown in \Cref{tab:bias_features}.

\begin{table}[h!]
\centering
\begin{tabular}{|c|c|l|c|}
\hline
\textbf{Value} & \textbf{Feature id} & \textbf{Feature description} & \textbf{Activation density} \\
\hline
-0.9743 & 1041 & Very large activations on the <BOS> token. & 42\% \\
\hline
0.8529 & 7507 & Large activations on the <BOS> token. & 48\% \\
\hline
-0.6471 & 2291 & No clear pattern. & 62\%  \\
\hline
-0.4107 & 12342 & Induction feature. & 27\% \\
\hline
-0.4065 & 7541 &  No clear pattern. & 42\% \\
\hline
-0.3745 & 1322 & No clear pattern. & 14\% \\
\hline
0.3522 & 2620 & All token positions > 2. & 98\% \\
\hline
0.3368 & 6810 & Mostly copyright licenses. & 45\% \\
\hline
0.3116 & 2514 & <BOS> tokens & 1\% \\
\hline
\end{tabular}
\caption{Top bias features.}
\label{tab:bias_features}
\end{table}

\begin{figure}[htbp]
  \centering
  \includegraphics[width=0.7\linewidth]{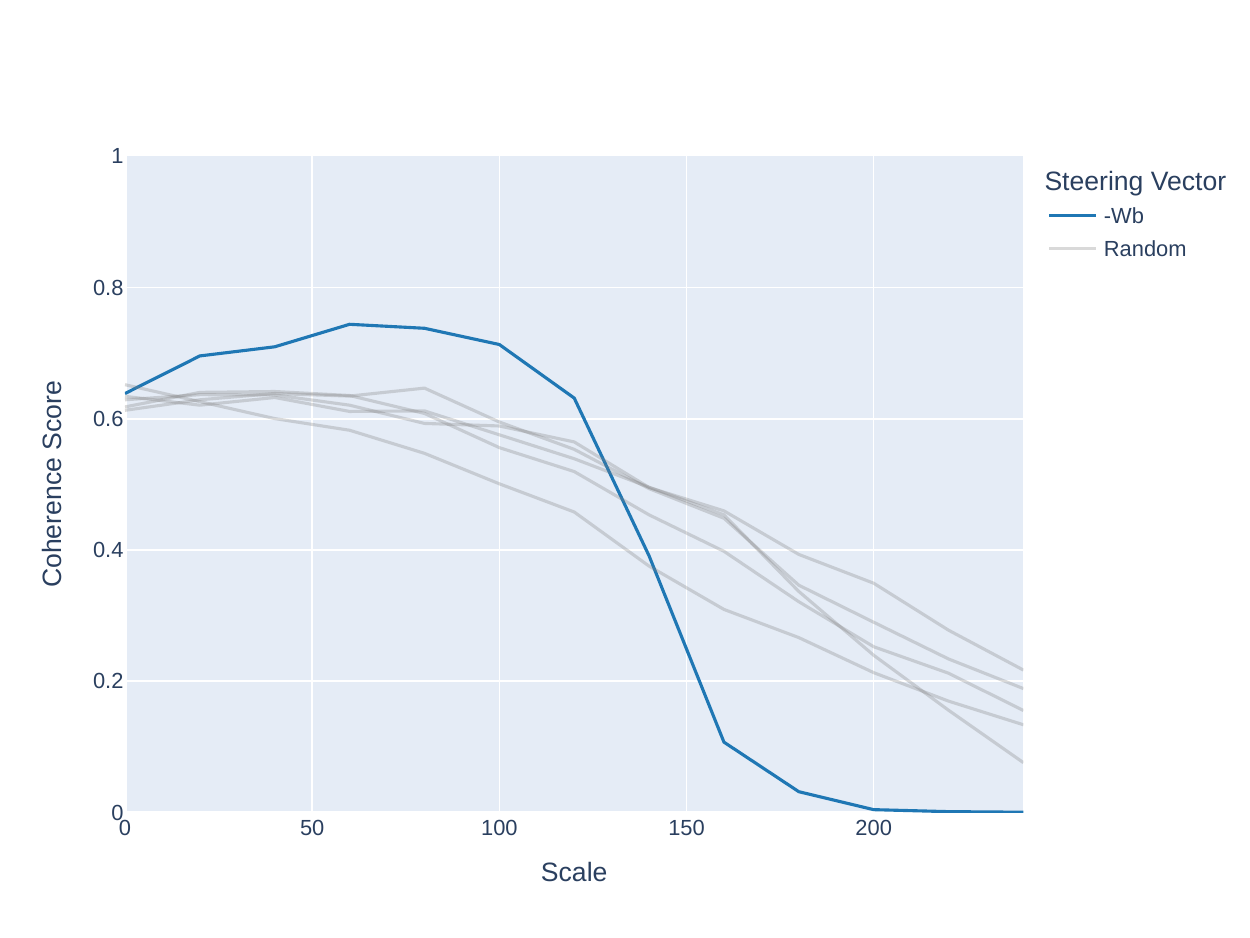}
  \caption{Coherence score when steering with the bias component of SAE-TS, compared to steering with random vectors. Steering with random vectors behaves as expected: Coherence score stays flat at first, then goes down gradually. However, when we steer with the bias term ($-\bm{Mb}$), Coherence score rises, peaking at 0.72 at scale 60, then drops off rapidly.}
  \label{fig:bias_coherence}
\end{figure}

We suspect that the initial Coherence bump that we observe in \Cref{sec:eval} is due to the bias term ($-\bm{Mb}$) in \Cref{eq:targeted_steering}. We can see this in \Cref{fig:bias_coherence} where we plot the Coherence score when steering with just $-\bm{Mb}$ against steering with random normal vectors.

One possible reason for the coherence bump could be due to some "incoherent text" features which on average activate more when we steer the model as the model is degraded. Then, when we subtract the bias term, the model produces more coherent outputs.

We notice some quirks when steering with the bias term -- notably, it seems to increase the frequency of proper nouns in the generated text, with almost every rollout containing a proper noun at scale 100. This raises the possibility that the apparent increase in coherence might be an artifact of our evaluation setup, where the presence of proper nouns causes gpt4o-mini to assign higher coherence scores rather than reflecting genuinely more coherent text. Example rollouts can be found in \Cref{subAppMinusMb}.

\section{Example Evaluation Criteria}
\label{sec:criteria}

We provide an example of our evaluation setup for the Wedding task:

\textbf{Behavioral score criterion:}
\begin{verbatim}
    The text contains discussion about a wedding
\end{verbatim}

\textbf{Coherence score criterion:}
\begin{verbatim}
    The text is coherent and the grammar is correct.
\end{verbatim}

\textbf{Evaluation Setup:}
We use gpt4o-mini with the following prompts:
\begin{tcolorbox}[colback=gray!10, colframe=black, boxrule=0.5pt]
\begin{verbatim}
System prompt

You score texts generated by a language model based on the following criterion:
[criterion].
You provide a score from 1 to 10. \
The language model was given a prompt and generated the following text. \
Evaluate the text based on the criterion. Output format should be \
JSON with the following fields: "score" (int)"
\end{verbatim}
\end{tcolorbox}

\begin{tcolorbox}[colback=gray!10, colframe=black, boxrule=0.5pt]
\begin{verbatim}
Message content

Prompt:

[prompt]

Completion:
"""
[completion]
"""
\end{verbatim}
\end{tcolorbox}

\section{Gemma-2-9B Results}
\label{sec:9B_results}

\begin{figure}[h!]
  \centering
  \includegraphics[width=\linewidth]{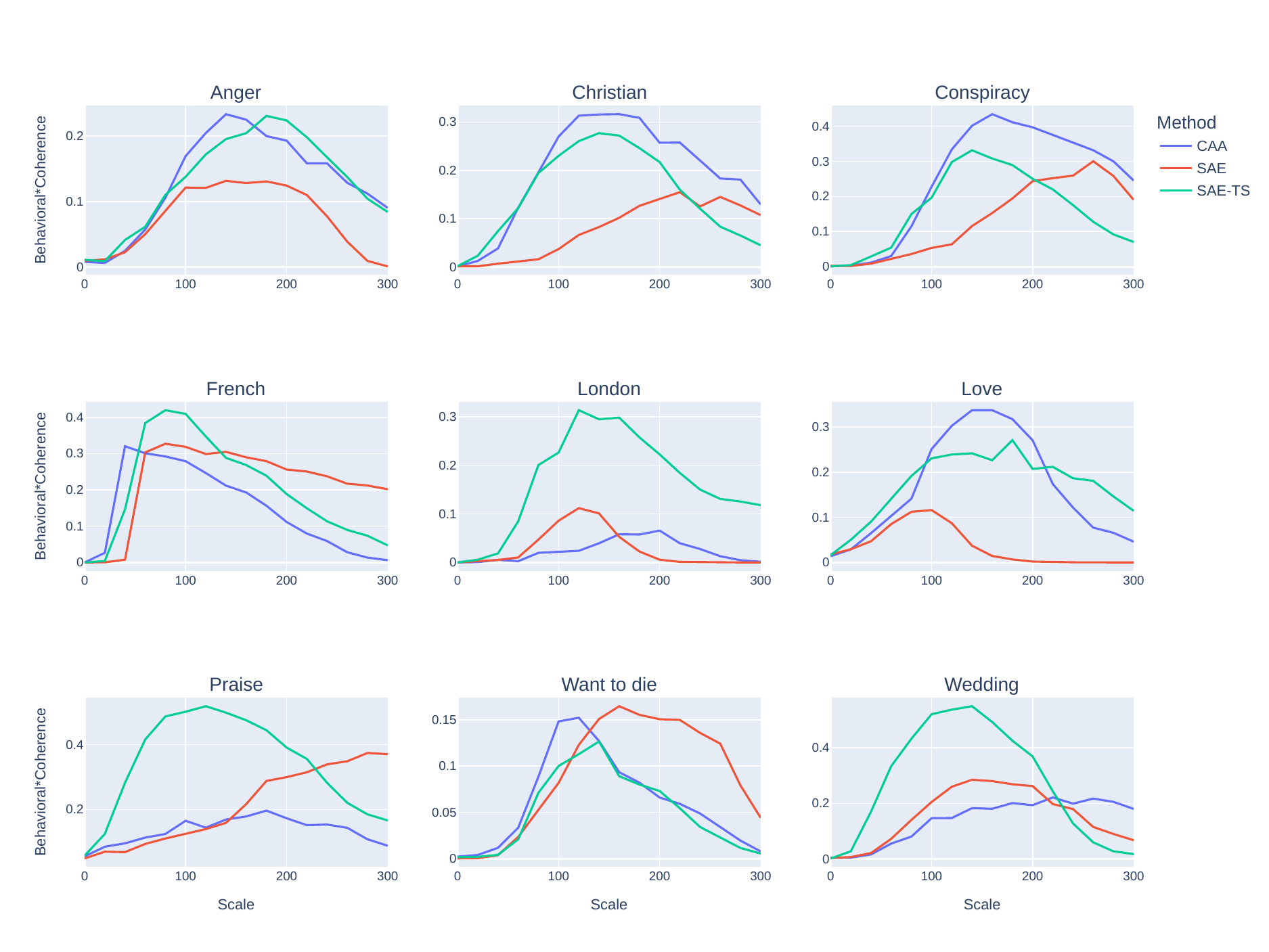}
  \caption{Gemma-2-9B plots showing Behavioral*Coherence score for the three methods over varying steering scales for all tasks.}
  \label{fig:all_plots_9B}
\end{figure}

We perform the same evaluations as in \Cref{sec:eval} for the gemma-2-9B model and plot the results in \Cref{fig:all_plots_9B} and \Cref{tab:9b_scores}. We use the Gemmascope layer 12 residual stream SAE with width 16k and an L0 of 130. It was easy to find relevant features for all the tasks except for the London task. There is no feature in this particular SAE which fires only on London, thus we use a mix of the following three features: \textbf{6915} - large cities including London, \textbf{9983} - cities, \textbf{11748} - United Kingdom. We see that SAE-TS achieves higher average Behavioral*Coherence score than either CAA or SAE steering. However, CAA beats SAE-TS in 4 of the 9 tasks and ties in one of them.

\begin{table}[h!]
\centering
\begin{tabular}{lccc}
\hline
\textbf{Steering Goal} & \textbf{CAA} & \textbf{SAE} & \textbf{SAE-TS (ours)} \\
\hline
Anger          & \textbf{0.2336} & 0.1317 & 0.2310 \\
Christian      & \textbf{0.3162} & 0.1548 & 0.2769 \\
Conspiracy     & \textbf{0.4343} & 0.3003 & 0.3315 \\
French         & 0.3203 & 0.3272 & \textbf{0.4196} \\
London         & 0.0656 & 0.1118 & \textbf{0.3135} \\
Love           & \textbf{0.3373} & 0.1160 & 0.2710 \\
Praise         & 0.1965 & 0.3743 & \textbf{0.5188} \\
Want to die    & 0.1521 & \textbf{0.1646} & 0.1265 \\
Wedding        & 0.2216 & 0.2851 & \textbf{0.5494} \\
\hline
Average        & 0.2531 & 0.2184 & \textbf{0.3376} \\
\hline
\end{tabular}
\caption{Gemma-2-9B maximum Behavioral*Coherence scores.}
\label{tab:9b_scores}
\end{table}

\section{Rotation Steering}
\label{sec:rotation_steer}
Our SAE-TS method requires measuring feature effects for each feature we want to target. However, some features might be too rare to appear in our measurement dataset, or might not exist in the SAE we use for measurement. To address this, we develop rotation steering, a method that learns a transformation $f: \mathbb{R}^{d_\mathrm{model}} \rightarrow \mathbb{R}^{d_\mathrm{model}}$ between SAE decoder vectors and effective steering vectors.

We compute this transformation by first calculating the correlation matrix $\bm{C} = \bm{M}\bm{W}_{\mathrm{dec}}$. We then perform SVD on this correlation matrix: $\bm{C} = \bm{U}\bm{\Sigma}\bm{V}^T$ and then the rotation matrix $\bm{R} = \bm{U}\bm{V}^T$ gives us a transformation between the model's activation spaces. Next, similarly to the process described in \Cref{sec:optimised_steering_vectors}, we generate a steering vector for any SAE feature $i$ by taking its decoder vector $\bm{d}_i$ (the $i$th row of $\bm{W}_{\mathrm{dec}}$) and apply the learned rotation to get $\bm{v} = \bm{d}_i\bm{R}$, we then normalise $\bm{v}$ and subtract (normalised) $\bm{M}\bm{b}$.

This approach allows us to generate steering vectors for features not present in our original feature effects dataset, as long as we can find a relevant feature in any SAE trained on the same model layer.

In \Cref{tab:rotation_steer_max_product} we see that rotation steering performs reasonably well but is on average worse than SAE-TS, at least for the specific tasks we test on.

\begin{table}[h!]
\centering
\begin{tabular}{lccccc}
\hline
\textbf{Steering Goal} & \textbf{CAA} & \textbf{SAE} & \textbf{SAE-TS} & \textbf{PinverseSteer} & \textbf{RotationSteer} \\
\hline
Anger          & 0.1666 & 0.0810 & 0.2424 & 0.1899 & \textbf{0.2460} \\
Christian      & \textbf{0.3486} & 0.0902 & 0.3302 & 0.0025 & 0.0870 \\
Conspiracy     & 0.3748 & 0.2319 & 0.3729 & 0.0644 & \textbf{0.4074} \\
French         & 0.2684 & 0.0589 & 0.3186 & 0.1912 & \textbf{0.3724} \\
London         & 0.0476 & 0.0061 & \textbf{0.5380} & 0.0722 & 0.2745 \\
Love           & 0.3039 & 0.1004 & \textbf{0.4234} & 0.2682 & 0.3851 \\
Praise         & 0.1332 & 0.2654 & \textbf{0.2842} & 0.0638 & 0.2833 \\
Want to die    & 0.1284 & 0.0649 & 0.1870 & 0.1683 & \textbf{0.2167} \\
Wedding        & 0.1768 & 0.2626 & \textbf{0.5432} & 0.0567 & 0.5313 \\
\hline
Average        & 0.2165 & 0.1290 & \textbf{0.3600} & 0.1197 & 0.3115 \\
\hline
\end{tabular}
\caption{Table showing maximum Behavioral*Coherence scores for various methods, including the three methods described in \Cref{sec:eval}, as well as finding the optimal steering vector by taking the pseudoinverse of $\bm{M}$, and rotation steering as described in \Cref{sec:rotation_steer}.}
\label{tab:rotation_steer_max_product}
\end{table}

\section{Finding related groups of features}
\label{sec: feature families}

We find that by looking at features with similar feature effects, we can find groups of features that are thematically related – more so than we can when looking at cosine similar features along their decoder vectors.

Process for curating the below example wedding feature group: Starting with the wedding feature, we recursively look at features with similar feature effects, adding the ones that we qualitatively believe fit into the feature group. We then repeat this process looking at features with similar decoder vectors. The venn diagram below compares the set of features that we could find looking at the SAE decoder vector vs. the feature effects.

This suggests that the SAE may not project features with similar effects into similar decoder directions. A research question to explore in the future would be to compare SAE decoder vectors from features in the same feature group to see if there are shared dimensions that encode important properties.

\begin{table}[h!]
\centering
\begin{tabular}{|c|c|c|}
\hline
\textbf{Cosine similarity} & \textbf{Feature effects} & \textbf{Both} \\
\hline
Parents & Wedding invitation  & Wedding \\
Siblings & Bride and groom  & Marriage \\
 & Ceremony  & Wife \\
 & Familial relationships  & Husband \\
 & Photography & Relationship \\
 & Venue &  \\
 & Party &  \\
 & Marriage act &  \\
 & Romantic  &  \\
 & In-laws  &  \\
 & Newly-wed  &  \\
\hline
\end{tabular}
\caption{Wedding features found by inspecting feature effect vs. SAE decoder vector similarities}
\label{tab:feature_families}
\end{table}

Note: We performed this experiment on the feature effects computed for all features in an SAE trained on activations from the 2B parameter Gemma 1 model early on in our research. However, we suspect the results here generalize to other LLM variants, as long as the feature effects map appropriately to observed model behaviors.

\clearpage
\section{Different Prompt}
\begin{figure}[h!]
  \centering
  \includegraphics[width=\linewidth]{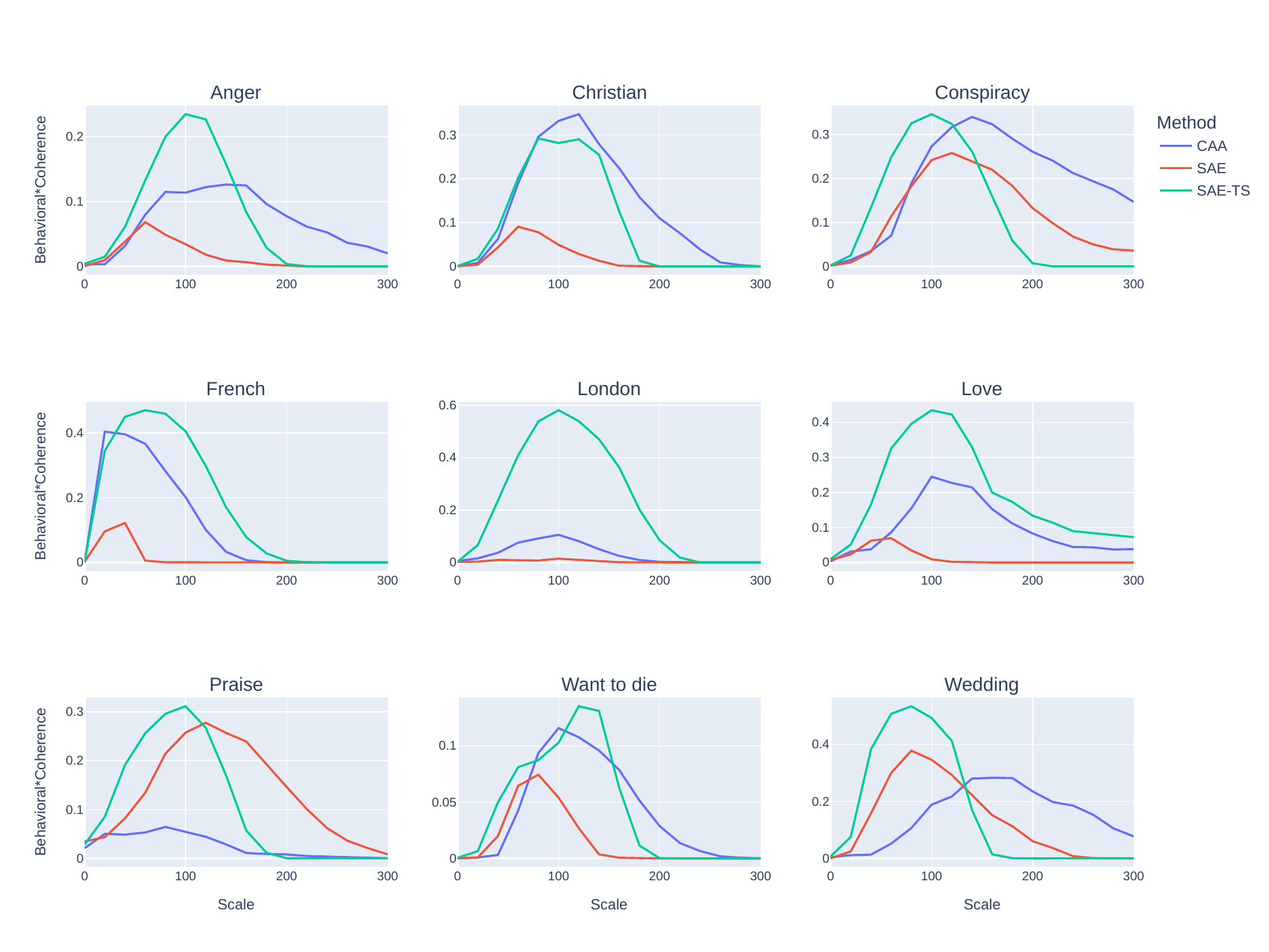}
  \caption{Plots showing evaluation scores for Gemma-2-2B starting from the prompt "<BOS>Surprisingly,".}
  \label{fig:all_plots_2B_surprisingly}
\end{figure}

Unless otherwise specified, all the steering score analysis in this paper was done with "<BOS>I think" as the starting prompt from which completions are generated. To verify that our results do not depend on this particular prompt, we repeat the evaluation experiments the "<BOS>Surprisingly," prompt (\Cref{fig:all_plots_2B_surprisingly}). We see that the results are qualitatively similar to the "<BOS>I think" prompt.

\clearpage
\section{Score Trade-off Curves}

\begin{figure}[htbp]
  \centering
  \includegraphics[width=\linewidth]{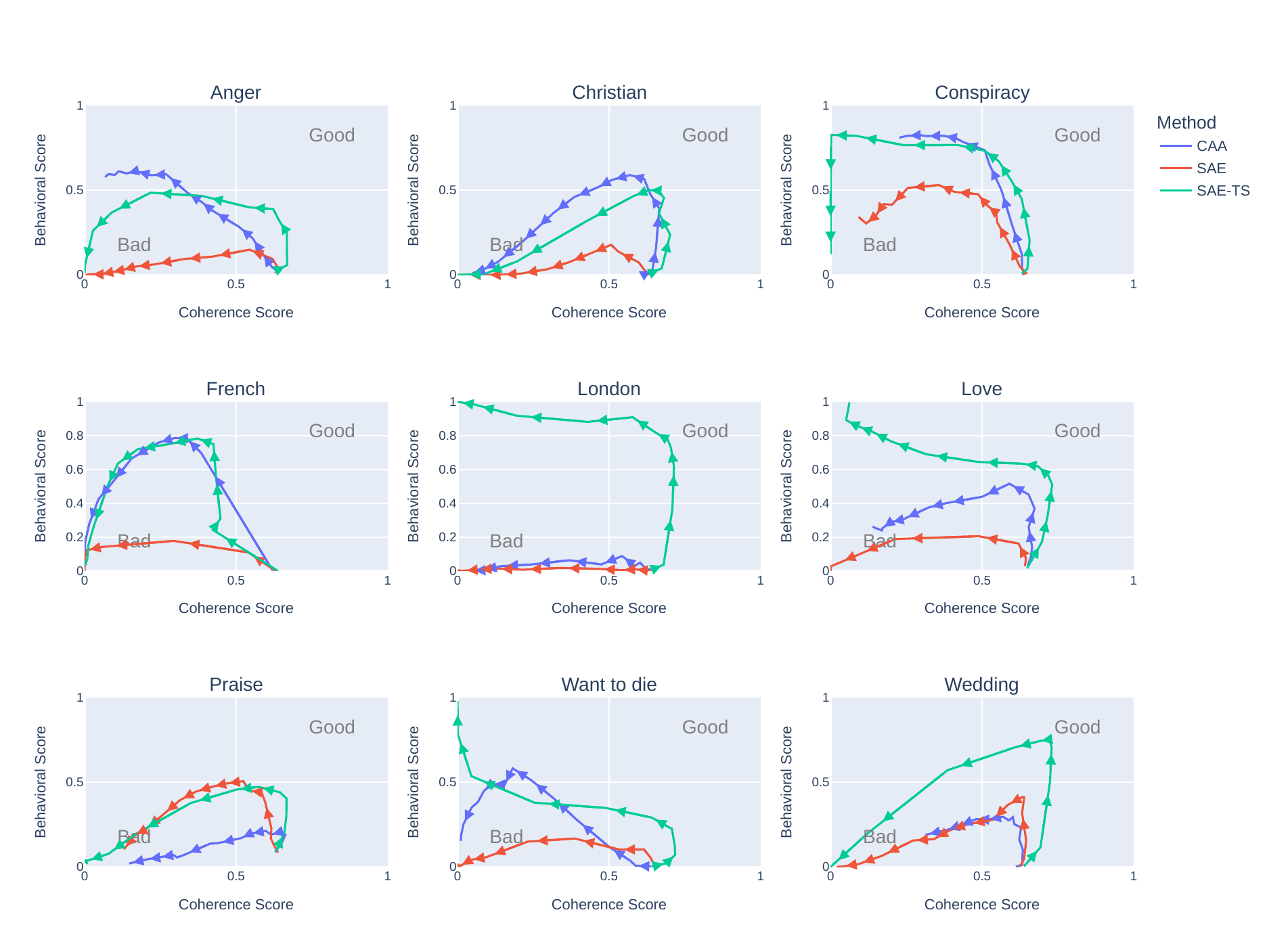}
  \caption{ Score trade-off curves. Each plot shows how Coherence score (x-axis) and Behavioral score (y-axis) change as the steering scale increases, with each line representing a different steering method. Points further to the right and top indicate better performance on both metrics.}
  \label{fig:parete_curves_2b}
\end{figure}

\clearpage
\section{Feature Effect Examples}
\label{sec:feature_effect_examples}

These are additional examples of feature effects that we find for various steering vectors extracted from Gemma-2-2b. Refer to \Cref{sec:measuring steering effects} for details on the method

\begin{table}[h!]
\centering
\begin{tabularx}{0.9\textwidth}{|c|c|X|}
\hline
\textbf{Activation} & \textbf{Feature id} & \textbf{Feature description} \\
\hline
2.4199 & 13243 & Wedding invitations and venue planning \\
\hline
2.3485 & 4230  & Wedding/marriage \\
\hline
1.0574 & 7507  & No clear pattern. Commonly occurring feature with 48\% activation density \\
\hline
0.8631 & 14599 & No clear pattern. Commonly occurring feature with 22\% activation density \\
\hline
0.7401 & 2421  & Marriage \\
\hline
0.7075 & 8522  & Visiting family \\
\hline
0.5743 & 11114 & Female family member \\
\hline
0.5212 & 11768 & Famous couples \\
\hline
0.5073 & 2144  & Family activities and hobbies \\
\hline
0.4646 & 13871 & Events and gatherings \\
\hline
\end{tabularx}
\caption{Table showing activations, feature ids, and feature descriptions of the wedding SAE feature steering vector (ft id 4230). Gemma-2-2b.}
\label{tab:feature_effects_wedding}
\end{table}

\begin{table}[h!]
\centering
\begin{tabularx}{0.9\textwidth}{|c|c|X|}
\hline
\textbf{Act diff} & \textbf{Feature id} & \textbf{Feature description} \\
\hline
1.5730 & 6810  & No clear pattern. Commonly occurring feature with 45\% activation density \\
\hline
1.4755 & 7507  & No clear pattern. Commonly occurring feature with 48\% activation density \\
\hline
0.8158 & 7972  & "Hello" \\
\hline
0.6514 & 16358 & The beginning of a sentence giving a plan or intentions \\
\hline
0.5458 & 3527  & Statements related to existential reflection \\
\hline
0.4950 & 6140  & Mathematical symbols and calculations \\
\hline
0.4810 & 7314  & Opening quotation mark to the start of a quote or speech \\
\hline
0.4604 & 9349  & Formalities involving physical gestures \\
\hline
0.4503 & 1768  & Questioning or reasoning about thought processes or experience \\
\hline
0.4498 & 2514  & Physical objects and structures \\
\hline
\end{tabularx}
\caption{Table showing activations, feature ids, and feature descriptions of the SAE feature steering vector (ft id 15874) in Gemma-2-2b, Layer 12. This feature activates on the token "say" said in various contexts}
\label{tab:feature_effects_say}
\end{table}

\clearpage
\section{Example Rollouts}
\label{sec:example_rollouts}

\subsection{SAE-TS London}
Scale = 100 (optimal scale)
\begin{verbatim}
<bos>I think these photos were taken at the National Portrait Gallery in London earlier 
    this year. The portraits collection features paintings of British actors Johnny Depp 
    and Helen Mirren as
<bos>I think I've written about my mum's beautiful family home in north London a few 
    times, but it was my last weekend in London with my mum
<bos>I think it's safe to say that the fashion scene in New York City continues to soar.
    \n\nAfter a few years of highlights from the city, including
<bos>I think we've all been waiting for a new <em>Sherlock</em> series over in London 
    ever since the actress Karen Wright announced that she would be leaving the
\end{verbatim}

Scale = 160 (too high). At high steering scales, the model seems to mostly talk about fashion, designers, and art exhibitions.
\begin{verbatim}
<bos>I thinkLondon designer Samantha Morton's kitchen installation recently designed 
    in London inspired by the city. The Interior Design show hosted event hosted in 
    London which saw Interiors Designer
<bos>I thinkyou artist  Victoria Morris's new exhibition at Somerset Gallery in London 
    is celebrating the French capital's fashion and design scene with Works on Display,
<bos>I thinkLondon's fashion scene is thriving with vibrant exhibits and exhibitions 
    currently showing in The Royal Academy's Summer Exhibition, 
    whichtookreturnsfromLondonfavouritesMichel",
<bos>I thinkLondon's fashion scene continues to thrive thanks to the city's eclectic 
    communities which celebrate the capital\u2019s vibrant eateries and restaurants 
    thissummer whichhave
\end{verbatim}

\subsection{CAA London}
Scale = 100.
\begin{verbatim}
<bos>I think they are a bit misleading as the term "London" as a hotel may be in poor
    taste.\n\nIf you look beyond that, it is a,
<bos>I think the biggest challenge this year for the FA has been what they do with the
    2nd stage teams. The clubs have been locked away from each other,
<bos>I think I read somewhere that they've been asked to be more involved in the music
    scene before. I mean I get that the band probably won't,
<bos>I think that people should feel 'theatred' rather than 'thespian' and 'theatre'
    rather than\n'thewest', 'South',
\end{verbatim}

\subsection{SAE London}
Scale = 120. Does not mention London.
\begin{verbatim}
"<bos>I think I have seen them on before. It is a very good idea.\n\nThanks, and 
    yes, it is a good idea..\n\nIf the market",
"<bos>I think this is 369.\n\nI' 60 in oner\n\nYeah we go and get him\n\nWell 
    done, now for those",
"<bos>I think you are still the same - the first time I met you to make my wife' 
    I know that it is no. 1 (I do not",
"<bos>I think that' 2015 is an actually a correct time to open this book, 
    as each of you will get in aneropolis after an",
\end{verbatim}

\subsection[Steering with -Mb]{Steering with $-\,\bm{M}\bm{b}$}
\label{subAppMinusMb}

Scale = 100.
\begin{verbatim}
"<bos>I think back on my childhood and remember growing up in Atlanta with my family 
    and best friend.\n\nWe spent our summers playing baseball in the neighborhood 
    and learning to"
"<bos>I think the idea of The Black Keys’ new video for “Lonely Phaser” is inspired 
    by a “Twilight” movie.\n\nShot in London and directed by"
"<bos>I think we’re officially in Fall. It’s been a chilly season for the past 
    couple months – and this week is no exception. While we’"
"<bos>I think the new movie “Still Alice” based on the memoir of Alice Koda is one 
    of the most compelling films of the year. The film follows the"
\end{verbatim}

\end{document}